\documentclass[journal]{IEEEtran}
\usepackage{cite}
\usepackage{times}
\usepackage{soul}
\usepackage{url}
\usepackage[hidelinks]{hyperref}
\usepackage[utf8]{inputenc}
\usepackage{amsmath}
\usepackage{booktabs}
\usepackage{algorithmic}
\usepackage{tabularx}
\usepackage[pdftex]{graphicx}
\ifCLASSINFOpdf
\else
\fi
\ifCLASSOPTIONcompsoc
 \usepackage[caption=false,font=normalsize,labelfont=sf,textfont=sf]{subfig}
\else
 \usepackage[caption=false,font=footnotesize]{subfig}
\fi

\begin{document}
%
\title{Learning Object Scale With Click Supervision for Object Detection}
%
%
%

\author{{Liao~Zhang,~Yan~Yan,~Lin~Cheng,~and~Hanzi~Wang$^\star$}
\thanks{L. Zhang, Y. Yan, L. Cheng and H. Wang are with Fujian Key Laboratory of Sensing and Computing for Smart City, School of Informatics, Xiamen University, Xiamen, 361005, China (e-mail:leochang@stu.xmu.edu.cn; cheng.charm.lin@hotmail.com; \{yanyan,Hanzi.Wang\}@xmu.edu.cn). $^\star$The corresponding author.}}

%
%

\markboth{Journal of \LaTeX\ Class Files,~Vol.~14, No.~8, August~2015}%
{Shell \MakeLowercase{\textit{et al.}}: Bare Demo of IEEEtran.cls for IEEE Journals}
%



\maketitle
\begin{abstract}
Weakly-supervised object detection has recently attracted increasing attention since it only requires image-level annotations. However, the performance obtained by existing methods is still far from being satisfactory compared with fully-supervised object detection methods. To achieve a good trade-off between annotation cost and object detection performance, we propose a simple yet effective method which incorporates CNN visualization with click supervision to generate the pseudo ground-truths (i.e., bounding boxes). These pseudo ground-truths can be used to train a fully-supervised detector. To estimate the object scale, we firstly adopt a proposal selection algorithm to preserve high-quality proposals, and then generate Class Activation Maps (CAMs) for these preserved proposals by the proposed CNN visualization algorithm called Spatial Attention CAM. Finally, we fuse these CAMs together to generate pseudo ground-truths and train a fully-supervised object detector with these ground-truths. 
Experimental results on the PASCAL VOC 2007 and VOC 2012 datasets show that the proposed method can obtain much higher accuracy for estimating the object scale, compared with the state-of-the-art image-level based methods and the center-click based method.
\end{abstract}
\footnote{\copyright 2019 IEEE. Personal use of this material is permitted.  Permission from IEEE must be obtained for all other uses, in any current or future media, including reprinting/republishing this material for advertising or promotional purposes, creating new collective works, for resale or redistribution to servers or lists, or reuse of any copyrighted component of this work in other works.}
\footnote{DOI: 10.1109/LSP.2019.2937387}
%
\IEEEpeerreviewmaketitle

\section{Introduction}
%
%
%
%
\IEEEPARstart{C}{onvolutional} Neural Network (CNN) has shown the superior capability to extract generic visual features from large-scale labeled datasets \cite{deng2009imagenet,coco}. By taking advantage of CNN, many state-of-the-art object detection methods \cite{ren2015faster,girshick2015fast,redmon2016you,liu2016ssd,mod} have obtained excellent performance based on the instance-level supervision. However, such a way usually requires expensive human efforts to collect the accurate bounding box annotations. To address this problem, some weakly-supervised detectors \cite{bilen2016weakly,jie2017deep,diba2017weakly,shen2018generative,wang2018collaborative,TS2C,pcl,Ge_2018_CVPR_multi_evidience,tang2018weaklyrpn} have been proposed and they only use image-level labels as supervision information to train the networks.
Although collecting a large number of image-level annotations (e.g., image retrieval using keywords) is much easier than collecting accurate bounding box annotations, the performance obtained by these weakly-supervised detectors is much worse than the fully-supervised detectors. 

In this paper, to effectively balance the annotation cost and the detection performance, we use the center click as supervision information, which can not only provide a strong anchor point for locating the object center but also greatly reduce the total annotation time compared with the bounding box annotation. According to the results in \cite{papadopoulos2017training}, manually annotating a high-quality bounding box needs 34.5 seconds on average, but annotating a high-quality center click just needs about 1.87 seconds. The time cost of annotating a center click is approximately equal to that of labeling an image (about 1.5 seconds).  The method in \cite{papadopoulos2017training} is the pioneering work to use the click supervision, which requires the annotators to click on the center of an imaginary bounding box that covers the object tightly and then incorporate the click supervision into a reference Multiple Instance Learning (MIL) framework. Compared with their method that use many refinement techniques to obtain competitive results, our method is much simpler while obtaining much better performance.

Recently, some CNN visualization methods \cite{zhou2016learningcam,selvaraju2017grad,li2018tellgain} have shown that the CNN model trained on the classification dataset has the remarkable localization ability. Zhou \emph{et al.}\cite{zhou2016learningcam} firstly generate the CAMs by accumulating a weighted sum of the last activation maps to visualize the class-specific regions. 
Selvaraju \emph{et al.}\cite{selvaraju2017grad} combine the activation maps with the corresponding gradients for a target class to obtain the CAMs, which can adapt to various off-the-shelf CNN architectures.

Inspired by these methods, we propose to estimate the object scale by generating the CAMs and then use pseudo ground-truths to train a fully-supervised detector.
However, when the training images contain the similar background or difficult instances, the CAMs obtained by these methods often cover background or highlight the discriminative and small regions. Therefore, simply using these CAMs to generate the pseudo ground-truths to train a fully-supervised detector will lead to poor performance.

To alleviate these problems, we propose to select proposals generated by the selective search algorithm \cite{uijlings2013selective}, instead of using the whole image, to train the CNN based classification network. To further improve the performance of weakly-supervised detectors, especially for detecting some small objects, we propose a pseudo ground truth generation method which incorporates a new CNN visualization algorithm called Spatial Attention CAM (SA-CAM) with center click. Specifically, the center click provides more information about object location, while CNN visualization can be used to estimate the object scale.
We summarize the main contributions of this paper as follows:
\begin{itemize}
\item We propose a simple yet effective method which combines a new CNN visualization algorithm called Spatial Attention CAM with center click to learn the object scale and  then generate accurate pseudo ground-truths for training a fully-supervised object detector, thus leading to significant performance improvements.
\begin{figure}[ht]
    \centering
    \includegraphics[width=1.0\linewidth,height=0.4\linewidth]{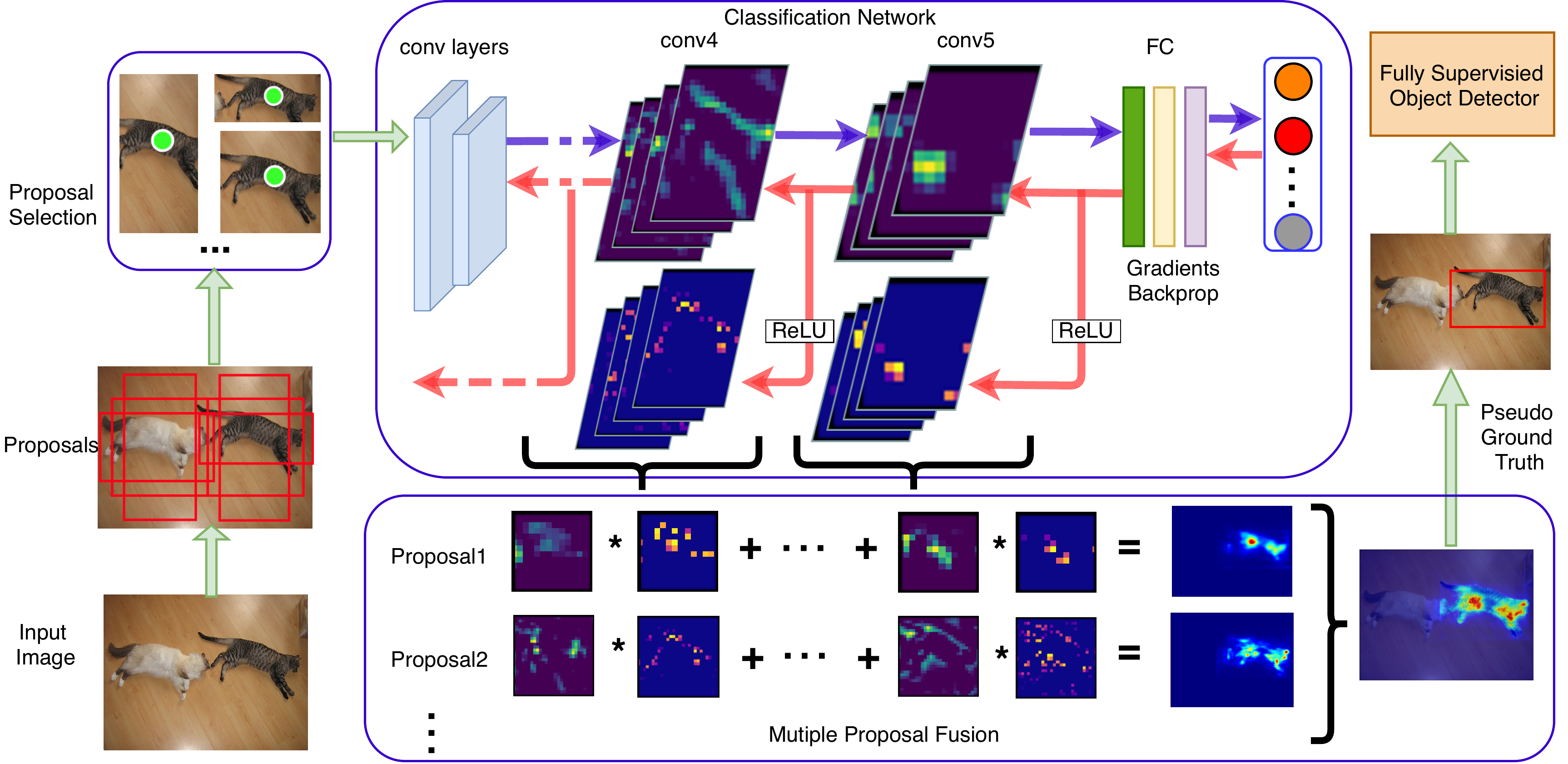}
    \caption{The network architecture of our method.}
    \label{fig:arch}
\end{figure}
\item Evaluations on the PASCAL VOC 2007 and VOC 2012 datasets show that our method outperforms several state-of-the-art weakly-supervised detection methods and a center-click based method by a large margin.
\end{itemize}

\section{The Proposed Method}
In this section, we introduce the proposed method in detail. Firstly, we select high-quality proposals generated by the selective search algorithm to train a classification network (Section II-A). Secondly, we generate the pseudo ground-truths by taking advantage of the proposed CNN visualization method (Section II-B). Finally, these pseudo ground-truths are used to train the fully-supervised object detectors (Section II-C). Fig. \ref{fig:arch} shows the network architecture of our method.
\subsection{Proposal Selection for Classification}
In this paper, we make use of center click as supervision information instead of only using image-level annotation. Hence, how to take advantage of center click to obtain tight and accurate pseudo ground-truths (considering different object scales), which can be used to train a fully-supervised object detector, is the main problem. Since the object detection datasets (e.g., PASCAL VOC) usually contain many challenging samples, simply applying these CNN visualization methods \cite{zhou2016learningcam, selvaraju2017grad} to estimate the object scale may fail.


Therefore, instead of training the classification network on the whole training image, we adopt a classification proposal selection algorithm to select proposals. For each image $I$, we firstly use the selective search algorithm \cite{uijlings2013selective} to generate about 2,000 proposals and then we filter out some proposals if there are no any center clicks inside them. For these preserved proposals, we rank them according to the Euclidean distance between the centers of these proposals and the positions of the center click. Note that these ranked proposals usually have big overlaps with each other. Therefore, we iteratively remove some highly overlapped proposals whose intersection-over-union (IoU) is larger than a threshold $T_{iou}$. Finally, we choose the top N proposals to train our classification network (in this paper, we use VGG16 \cite{simonyan2014verydeep} pretrained on ImageNet \cite{deng2009imagenet} as our backbone). 
Since these selected proposals contain less background and they contain more details, we can estimate the object scale more accurately.

\subsection{SA-CAM for Pseudo Bounding Box Generation}
Based on the trained classification network, we can estimate the object scale via some existing CNN visualization methods (e.g., \cite{zhou2016learningcam},\cite{selvaraju2017grad}). However, these methods usually lead to poor performance. In this paper, inspired by the Grad-CAM \cite{selvaraju2017grad}, we propose the  Spatial Attention CAM, namely SA-CAM, to better estimate the object scale. 

Given an image $I$, let $A_{n,l,k}$ be the $k$-th activation map in the $l$-th layer of the $n$-th proposal. For a target class c, how to discriminate each pixel of $A_{n,l,k}$ corresponding to c from other classes or background is the main issue for estimating the object scale. Grad-CAM firstly computes the gradients of the score $y^{c}$ (which is the output of the last fully-connected layer corresponding to class c), with respect to the activation map $A_{n,l,k}$, (i.e.\ $ \frac{\partial{y^c}}{\partial {A_{n,l,k}}}$). And then, these gradients are global-average-pooled to obtain the weight $w^{c}_{n,l,k}$ of a neuron:
\begin{equation}
\label{eqn1}
w^{c}_{n,l,k} = GAP \left ( \frac{\partial{y^c}}{\partial {A_{n,l,k}}} \right)
\end{equation}
where $GAP \left (\cdot \right)$ is the global average pooling operation.

Finally, the $l$-th layer CAM $M_{n,l}^{c}$ can be generated based on a weighted sum of activation maps and a ReLU operation.
\begin{equation}
\label{eqn2}
M_{n,l}^{c} = ReLU\left (  \sum_{k} w_{n,l,k}^{c}A_{n,l,k}\right )
\end{equation}

Grad-CAM uses GAP to obtain the average value of gradients (which can be positive or negative) as the weight for each activation map. However, it may not be precise enough to generate the CAMs by using a weighted sum of activation maps (because that only the positive gradients of each activation map indicate the presence of the target class c). Therefore, we firstly use a ReLU function to preserve those positive gradients and remove the negative gradients. Then, we propose to assign an attention score, which indicates the importance of each pixel belonging to the target class $c$, for each pixel of $A_{n,l,k}$ by multiplying the gradients with the corresponding pixel values. Specifically, the proposed SA-CAM is computed as:

\begin{equation}
\label{eqn3}
    \tilde{M}_{n,l}^{c} = \sum_k ReLU\left ( \frac{\partial{y^c}}{\partial {A_{n,l,k}}} \right ) A_{n,l,k}
\end{equation}

Based on Eq. (\ref{eqn3}), we firstly generate the CAMs for the selected proposals. And then, the generated CAMs are upsampled into their original sizes because these selected proposals have different sizes. Furthermore, we project these CAMs back to their original positions in the image $I$. Finally, these CAMs are fused together to get a higher resolution.
\begin{equation}
\label{eqn4}
    \tilde{M}_{fused}^{c} =  \sum_n F\left( \tilde{M}_{n,l}^{c}\right)
\end{equation}where $F \left (\cdot \right)$ is the function that upsamples the CAMs to their original sizes and then projects them back to their original positions in the image $I$.

After obtaining the fused CAMs, we can easily get the discriminative regions by using a threshold $T_{cam}$. However, these regions are not accurate enough to generate the pseudo ground-truths for training a fully-supervised object detector. Note that the center click provides a strong anchor point for the object center location. Therefore, we can get more accurate centro-symmetric regions by utilizing these center clicks. These centro-symmetric regions can be used to generate the pseudo ground-truths.
\subsection{Fully-Supervised Detector}
For improving the performance of weakly-supervised object detectors, we propose to use these generated pseudo ground-truths to train a fully-supervised detector. Considering both accuracy and speed, we adopt the Fast-RCNN based on VGG16 as the fully-supervised detector. 
\section{Experiments}

\subsection{Datasets and Evaluation Metrics}

\textbf{Datasets.} We conduct the experiments on the PASCAL VOC 2007 and VOC 2012 datasets \cite{everingham2010pascal}, which are 
commonly used for the weakly-supervised objection detection task. We use the center click annotation data provided by \cite{papadopoulos2017training}.

\textbf{Evaluation Metrics.} Mean Average Precision (mAP) is the standard metric used in PASCAL VOC, which evaluates the performance of an object detector. Correct Localization (CorLoc) \cite{deselaers2012weakly} calculates the percentage of the bounding boxes (generated by the evaluated algorithm) that correctly localize an object of the target class. Both of the two metrics consider a bounding box as a true positive sample if its IoU with the ground-truth target region is larger than 0.5.

\textbf{Implementation Details.} We conduct our experiments based on Caffe \cite{jia2014caffe} and use VGG16 pretrained on ImageNet as the classification backbone network. 
We set $T_{iou}$ to 0.7 for removing the overlapped proposals and select 8 proposals (i.e.\ $ N$ = 8) for each center click.
For generating pseudo ground-truths, we experimentally set $T_{cam}$ to be 80 if the number of image labels is less than 2 or set it to be 120 if the number is larger than 2. We follow the training strategy of Fast-RCNN \cite{girshick2015fast} to train a fully-supervised detector. Our experiments are conducted on a NVIDIA Titan XP GPU.
\subsection{Comparison with State-of-the-Art Methods}
We compare our method with several state-of-the-art image-level based detection methods \cite{bilen2016weakly,jie2017deep,diba2017weakly,shen2018generative,wang2018collaborative,TS2C,pcl,Ge_2018_CVPR_multi_evidience,tang2018weaklyrpn} and a center-click (one click) based method \cite{papadopoulos2017training}. 
Table \ref{tab:cor_loc_voc07} shows the CorLoc performance evaluated on the PASCAL VOC2007 trainval set. Bold font indicates best results obtained by the competing methods. Our method outperforms all the image-level based detection methods by a significant margin (11.2\%-26.1\%). Moreover, it also outperforms  the center-click based method by 6.3\% and obtains better CorLoc performance for most of the classes, especially for small objects (e.g., ``boat", ``bottle", ``plant"). The main reason is that our method can select high-quality proposals and thus it can estimate the object scale accurately. As a result, our method can obtain higher mAP performance than the competing methods. Table \ref{tab:map_voc07} shows the mAP performance on the PASCAL VOC2007 test set.  Note that except for \cite{bilen2016weakly}, the results obtained by the competing methods are achieved via training fully supervised object detectors. Our method trained on VGG16 achieves 56.8\% mAP, outperforming all the state-of-the-art image-level based methods, which also use VGG16 as backbone, by at least 5.4\%. Note that the center-click based method in \cite{papadopoulos2017training} uses AlexNet\cite{krizhevsky2012imagenet} as the backbone. For fair comparison, we replace VGG16 with AlexNet as done in \cite{papadopoulos2017training}. As a result, our method achieves 47.6\% mAP, which is better than 45.9\% mAP obtained by the method in \cite{papadopoulos2017training}.

We also conduct the experiments on the PASCAL VOC 2012 set. As shown in Table \ref{tab:corloc_voc12}, our method achieves 78.2\% CorLoc and 54.6\% mAP, which still outperforms all the image-level based methods and the center-click based method. The mAP results can be found at:\href{http://host.robots.ox.ac.uk:8080/anonymous/2PLONT.html}{http://host.robots.ox.ac.uk:8080/anonymous/2PLONT.html}
\subsection{Ablation Studies}
In this subsection, we firstly conduct an ablation experiment to show the influence of using different types of weak information on the performance of the proposed method. Then, we study the influence of back-propagating gradients to different layers on the performance of the proposed method. Finally, we also validate the effectiveness of the proposed SA-CAM.

\textbf{Influence of Using Different Types of Weak Information.}
Table \ref{tab:compare_cor_map} shows the comparison results obtained by the proposed method using different types of weak information (i.e., image labels and one center click). From the results, we can see that only using image labels to estimate the object scale lead to poor performance (60.5\% CorLoc and 46.7\% mAP). In contrast, our one-click based method achieves 79.6\% CorLoc and 56.8\% mAP, which achieves 19.1\% improvement on CorLoc and about 10.1\% improvement on mAP compared with our method only using image-label annotation. Therefore, center click is an efficient annotating strategy for improving the performance of weakly supervised object detectors. 

\textbf{Influence of Back-Propagating Gradients to Different Convoluational Layers.}
Table \ref{tab:best_conv} shows the CorLoc results obtained by the proposed method through back-propagating gradients to different convolutional layers on the PASCAL VOC 2007 trainval set. As shown in Table \ref{tab:best_conv}, 
back-propagating gradients to the conv5\_2 layer of VGG16 can obtain better CorLoc performance than back-propagating gradients to a lower layer (e.g., the conv4\_3 layer) or a higher layer (e.g., the conv5\_3 layer). This is because that back-propagating gradients to a lower layer may lead to unreliable results, while back-propagating gradients to a  higher layer may provide more semantic information but less spatial details, leading to poor performance. Therefore, back-propagating gradients to the middle layer (i.e., the conv5\_2 layer) makes a good trade-off and thus obtains the best results. We also conduct the experiments to show if combining the gradients from higher-level feature maps (e.g., the conv5\_2 layer) and lower-level feature maps (e.g., the conv5\_1 layer) can further improve the performance. 
However, combining the gradients from the conv5\_2 layer and the gradients from conv5\_1 layer achieves 75.3\% CorLoc, which is lower than only using the conv5\_2 layer (79.6\% CorLoc). 
\begin{table*}[!htb]
\centering
\caption{Comparison of Our Method with Ten State-of-the-art Methods on the PASCAL VOC2007 Trainval Set in Terms of CorLoc(\%).}
\label{tab:cor_loc_voc07}
\resizebox{\textwidth}{!}{
\begin{tabular}{lrrrrrrrrrrrrrrrrrrrrrr}
\hline
Method  & aero & bike& bird & boat & bottle & bus & car & cat & chair & cow & table & dog & horse & mbike &person & plant & sheep  & sofa & train & tv & CorLoc \\
\hline
Bilen \emph{et al.} \cite{bilen2016weakly}&65.1 &58.8 &58.5 &33.1 &39.8 &68.3 &60.2 &59.6 &34.8 &64.5 &30.5 &43.0 &56.8 &82.4 &25.5 &41.6 &61.5 &55.9 &65.9 &63.7 &53.5\\
\hline
Jie  \emph{et al.} \cite{jie2017deep} &72.7 &55.3 &53.0 &27.8 &35.2 &68.6 &81.9 &60.7 &11.6 &71.6 &29.7 &54.3 &64.3 &88.2 &22.2 &53.7 &72.2 &52.6 &68.9 &75.5 &56.1 \\
Diba \emph{et al.} \cite{diba2017weakly} &83.9 &72.8 &64.5 &44.1 &40.1 &65.7 &82.5 &58.9 &33.7 &72.5 &25.6 &53.7 &67.4 &77.4 &26.8 &49.1 &68.1 &27.9 &64.5 &55.7 &56.7 \\
Wei \emph{et al.}\cite{TS2C} &84.2 &74.1 &61.3 &52.1 &32.1 &76.7 &82.9 &66.6 &42.3 &70.6 &39.5 &57.0 &61.2 &88.4 &9.3 &54.6 &72.2 &60.0 &65.0 &70.3 &61.0\\
Wang \emph{et al.} \cite{wang2018collaborative}&85.8 &80.4 &73.0 &42.6 &36.6 &79.7 &82.8 &66.0 &34.1 &78.1 &36.9 &68.6 &72.4 &91.6 &22.2 &51.3 &79.4 &63.7 &74.5 &74.6 &64.7 \\
Tang \emph{et al.} \cite{pcl} &83.8 &\textbf{85.1} &65.5 &43.1 &50.8 &\textbf{83.2} &85.3 &59.3 &28.5 &82.2 &57.4 &50.7 &85.0 &92.0 &27.9 &54.2 &72.2 &65.9 &77.6 &\textbf{82.1} &66.6\\
Ge \emph{et al.}\cite{Ge_2018_CVPR_multi_evidience}&\textbf{88.3} &77.6 &74.8 &63.3 &37.8 &78.2 &83.6 &72.7 &19.4 &79.5 &46.4 &78.1 &84.7 &90.4 &28.6 &43.6 &76.3 &68.3 &77.9 &70.6 &67.0\\
Shen \emph{et al.} \cite{shen2018generative} & 76.5 &76.1 &64.2 &48.1 &52.5 &80.7 &86.1 &73.9 &30.8 &78.7 &62.0 &71.5 &46.7 &86.1 &60.7 &47.8 &82.3 &74.7 &\textbf{83.1} &79.3 &68.1\\
Tang \emph{et al.} \cite{tang2018weaklyrpn} &83.8 &82.7 &60.7 &35.1 &53.8 &82.7 &\textbf{88.6} &67.4 &22.0 &86.3 &68.8 &50.9 &\textbf{90.8} &\textbf{93.6} &44.0 &61.2 &82.5 &65.9 &71.1 &76.7 &68.4\\
\hline
Dim \emph{et al.} \cite{papadopoulos2017training}&- &- &- &- &- &- &- &- &- &- &- &- &- &- &- &- &- &- &- &- &73.3\\
Ours&87.5 &80.8 &\textbf{87.1} &\textbf{63.8} &\textbf{59.2} &78.2 &83.3 &\textbf{85.8} &\textbf{73.1} &\textbf{89.7} &\textbf{72.2} &\textbf{78.6} &90.5 &87.6 &\textbf{77.7} &\textbf{72.5} &\textbf{86.6} &\textbf{79.0} &82.5 &76.7 &\textbf{79.6} \\
\hline
\end{tabular}}
\end{table*}
The main reason is that the back-propagated gradients should match the corresponding pixels of feature maps.

\begin{table}[!htb]
\centering
\caption{Comparison of Our Method (using VGG16 or AlexNet) with Ten State-of-the-art Methods on the PASCAL VOC2007 Test Set in Terms of mAP(\%).}
\label{tab:map_voc07}
\begin{tabular}{lllr}
\hline
Method  &Backbone &Detector& mAP\\ 
\hline
Bilen \emph{et al.} \cite{bilen2016weakly} &VGG16&WSDNN\cite{bilen2016weakly}&39.5\\
\hline
Jie \emph{et al.} \cite{jie2017deep}&VGG16 &Fast-RCNN&41.7\\
Diba \emph{et al.} \cite{diba2017weakly}&VGG16 &Fast-RCNN&42.8\\
Shen \emph{et al.} \cite{shen2018generative}&VGG16 &SSD\cite{liu2016ssd}&47.0\\
Wei \emph{et al.}\cite{TS2C}&VGG16 &Fast-RCNN &48.0\\
Wang \emph{et al.} \cite{wang2018collaborative}&VGG16 &Faster-RCNN& 48.3\\
Tang \emph{et al.} \cite{pcl}&VGG16 &Fast-RCNN&48.8\\
Tang \emph{et al.} \cite{tang2018weaklyrpn}&VGG16 &Fast-RCNN&50.4\\
Ge \emph{et al.}\cite{Ge_2018_CVPR_multi_evidience}&VGG16&Fast-RCNN&51.2\\
\hline
Dim \emph{et al.} \cite{papadopoulos2017training}&AlexNet&Fast-RCNN &45.9\\
Ours &AlexNet&Fast-RCNN& 47.6\\
Ours &VGG16&Fast-RCNN&\textbf{56.8}\\
\hline
\end{tabular}
\end{table}
\begin{table}[!htb]
\centering
\caption{Comparison of Our Method to Six State-of-the-art Methods on the PASCAL VOC2012 Data Set in Terms of CorLoc (\%) And mAP (\%).}
\label{tab:corloc_voc12}
\begin{tabular}{llllr}
\hline
Method &CorLoc&Backbone&Detector&  mAP \\
\hline
Wang \emph{et al.} \cite{wang2018collaborative} &65.2&VGG16&Faster-RCNN&43.3\\
Tang \emph{et al.} \cite{pcl} &68.0&VGG16&Fast-RCNN&44.2\\
Wei \emph{et al.}\cite{TS2C}&64.4&VGG16&Fast-RCNN&44.4\\
Tang \emph{et al.} \cite{tang2018weaklyrpn} &69.3&VGG16&Fast-RCNN&45.7\\
Ge \emph{et al.}\cite{Ge_2018_CVPR_multi_evidience}&69.4&VGG16&Fast-RCNN&47.5\\
\hline
Dim \emph{et al.} \cite{papadopoulos2017training} &67.2&-&-&-\\
Ours &\textbf{78.2} &VGG16&Fast-RCNN&\textbf{54.6}\\
\hline
\end{tabular}
\end{table}
\begin{table}[!ht]
\centering
\caption{CorLoc (\%) and mAP (\%) obtained by the Proposed Method Using Two Different Types of Weak Information on the Pascal VOC2007 Data Set.}
\label{tab:compare_cor_map}
\begin{tabular}{lllc}
\hline
Weak Information &Detector&CorLoc & mAP \\
\hline
image label &Fast-RCNN& 60.5 & 46.7\\
one-click &Fast-RCNN& \textbf{79.6} & \textbf{56.8}\\
\hline
\end{tabular}
\end{table}
\begin{table}[!ht]
\centering
\caption{CorLoc (\%) Obtained by the Proposed Method About Back-Propagating Gradients to Different Convoluational Layers on the PASCAL VOC2007 Trainval Set.}
\label{tab:best_conv}
\begin{tabular}{llc}
\hline
Convoluational layer &Gradients layer & CorLoc \\
\hline
conv4\_3&conv4\_3&72.6 \\
conv5\_1&conv5\_1&77.0 \\
conv5\_2&conv5\_2&\textbf{79.6} \\
conv5\_3&conv5\_3&76.4\\
conv5\_1&conv5\_2&75.3 \\
\hline
\end{tabular}
\end{table}

\begin{figure}[!htb]
\centering
\subfloat{\includegraphics[width=0.12\textwidth,height=0.1\textwidth]{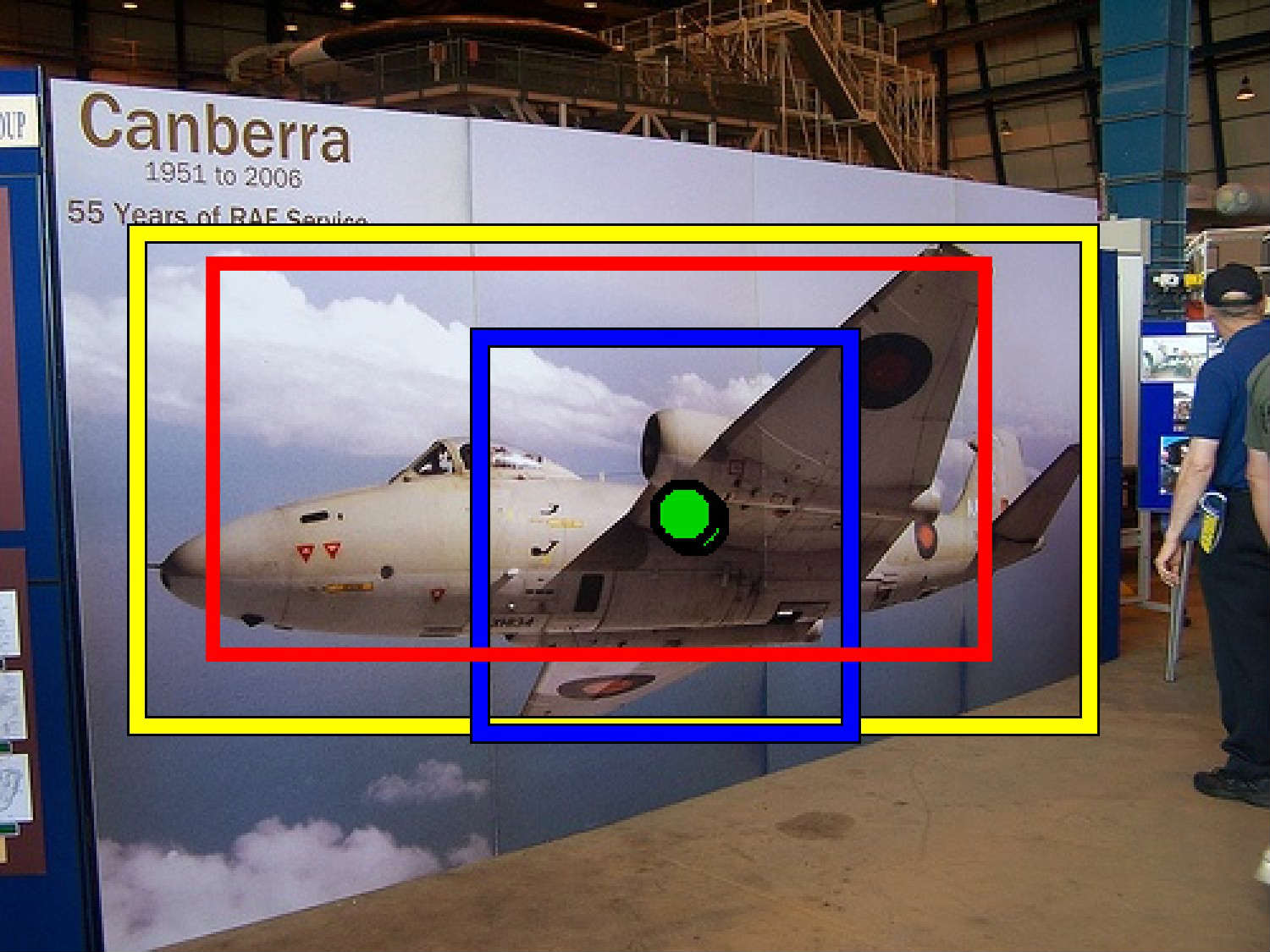}}\hfill
\subfloat{\includegraphics[width=0.12\textwidth,height=0.1\textwidth]{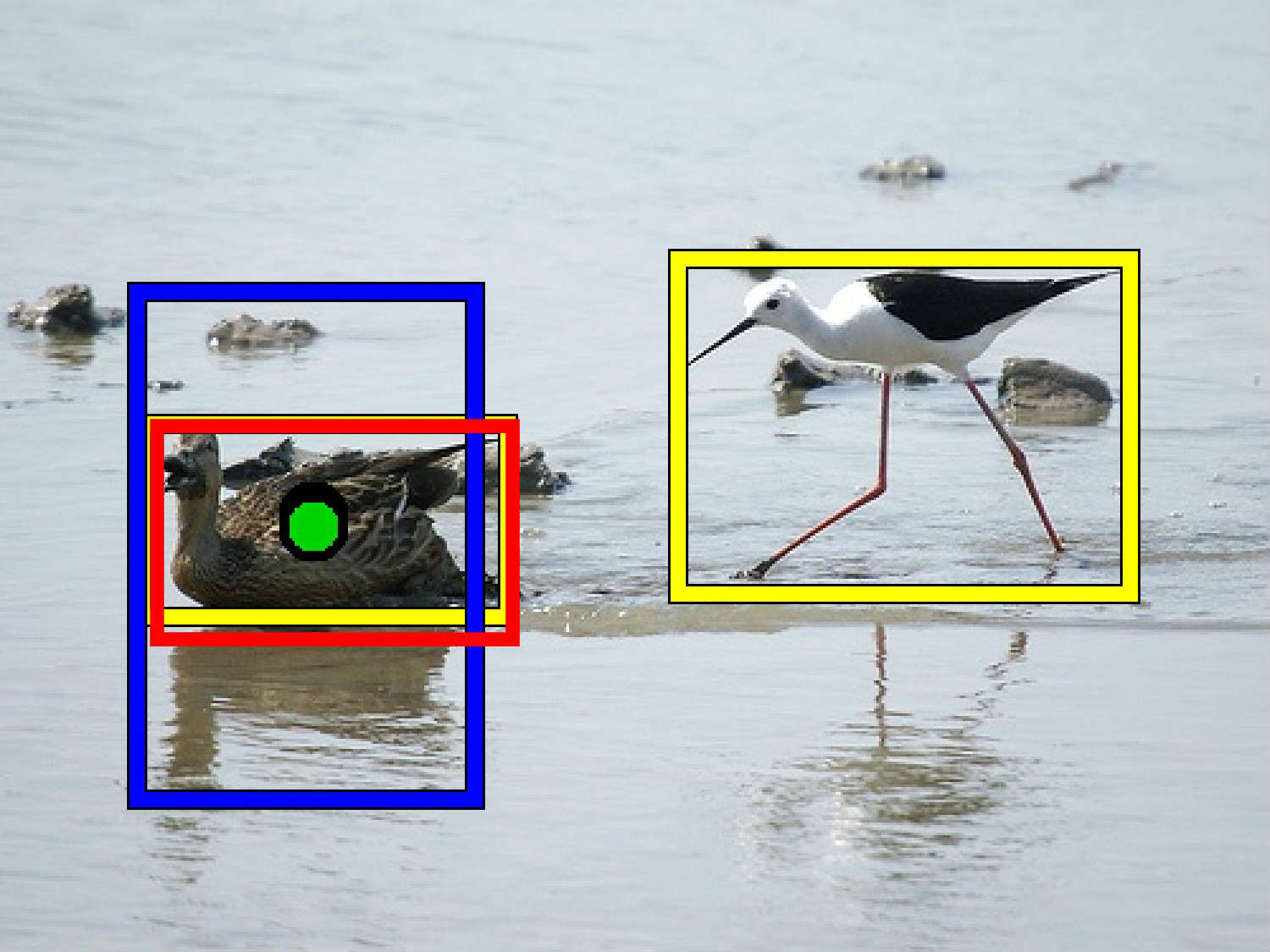}}\hfill
\subfloat{\includegraphics[width=0.12\textwidth,height=0.1\textwidth]{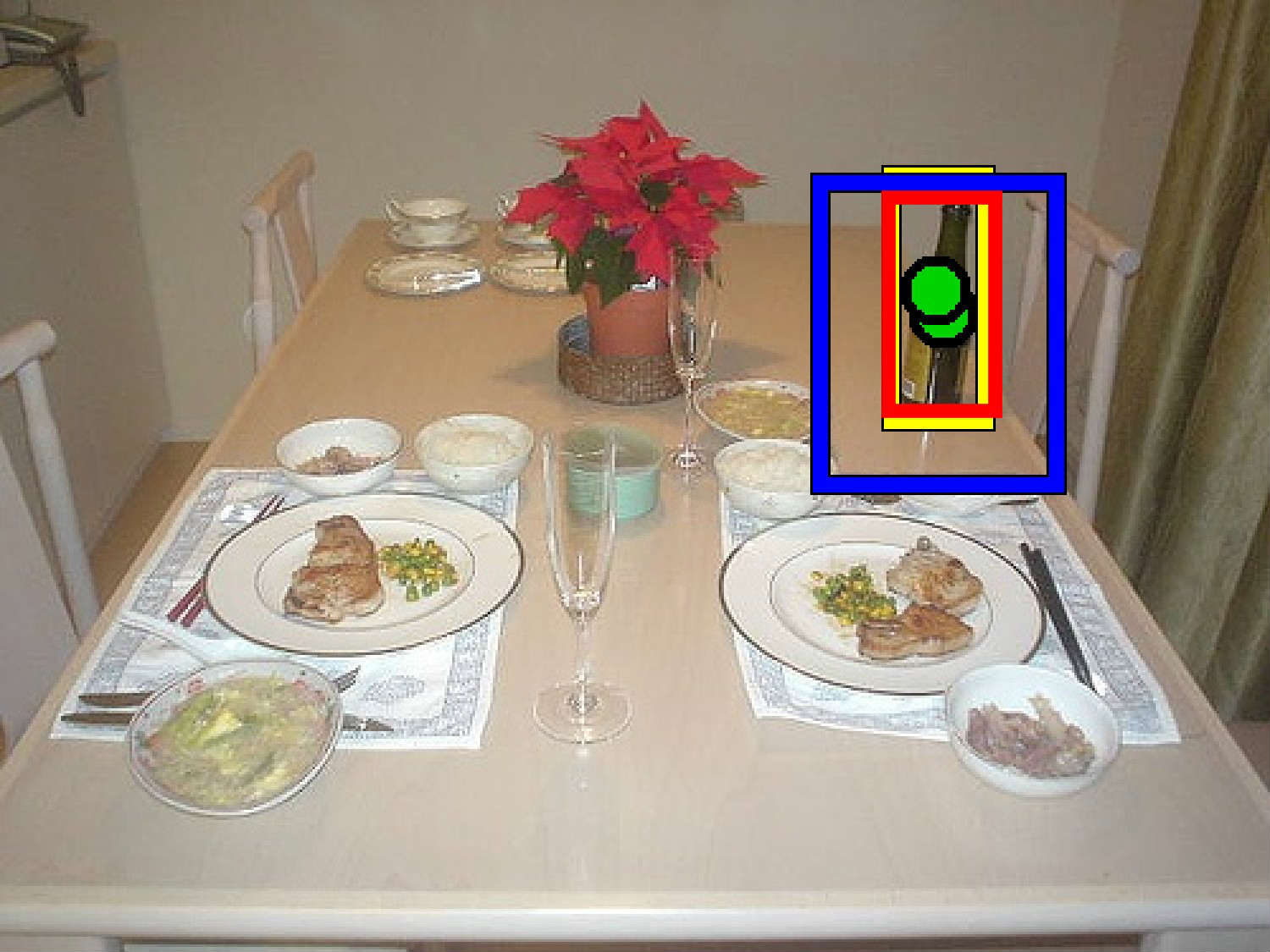}}\hfill
\subfloat{\includegraphics[width=0.12\textwidth,height=0.1\textwidth]{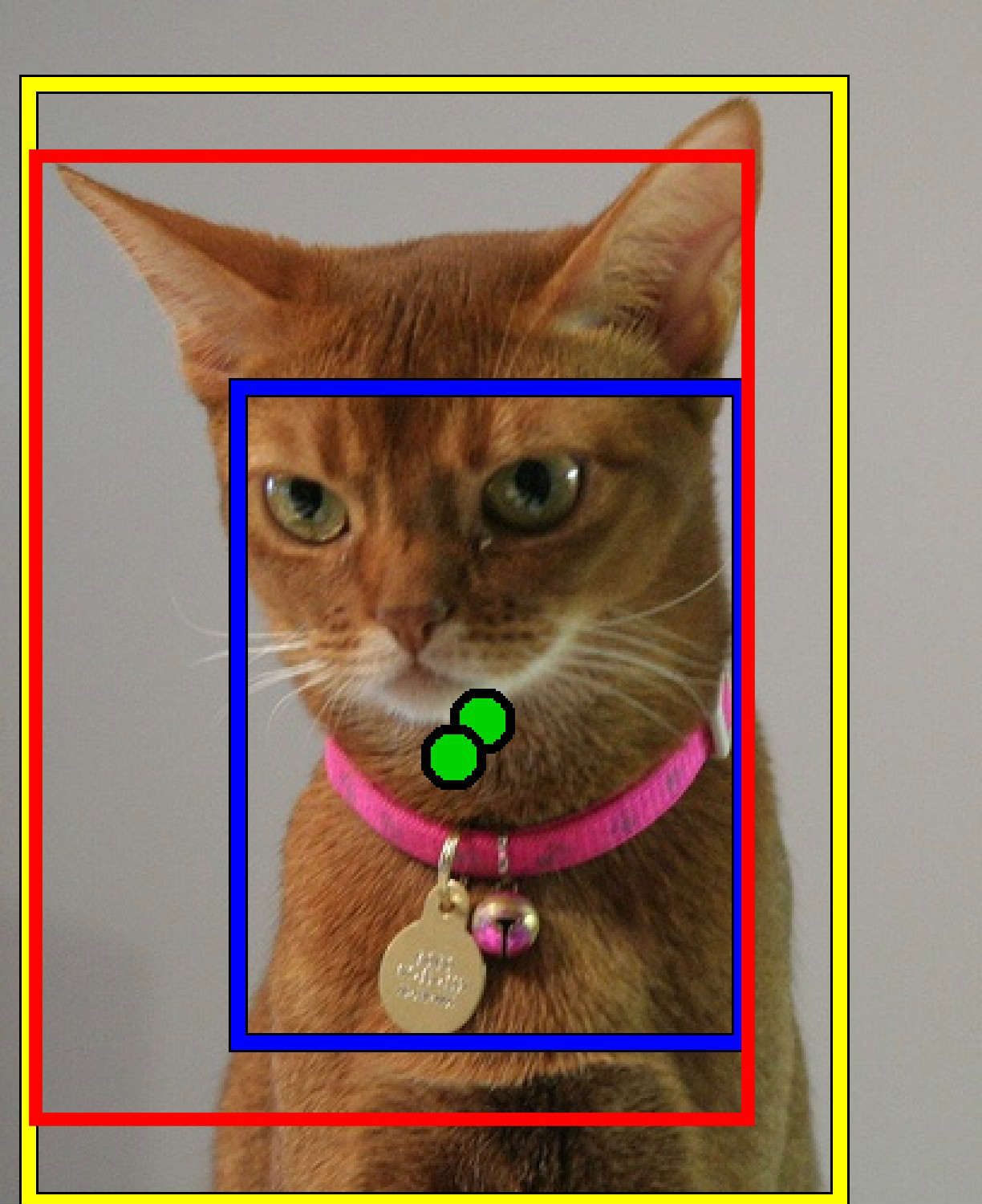}}\hfill
\subfloat{\includegraphics[width=0.12\textwidth,height=0.1\textwidth]{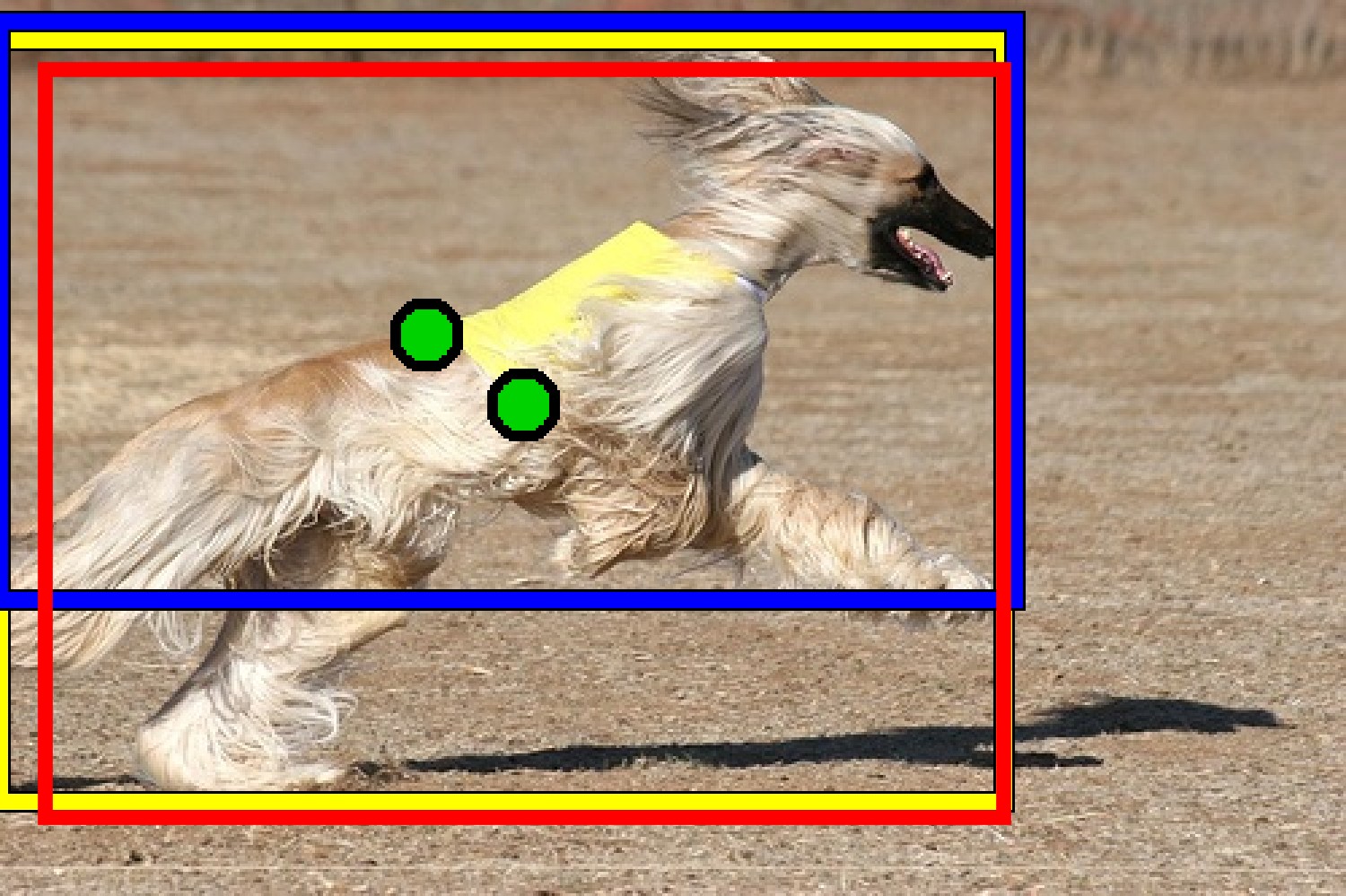}}\hfill
\subfloat{\includegraphics[width=0.12\textwidth,height=0.1\textwidth]{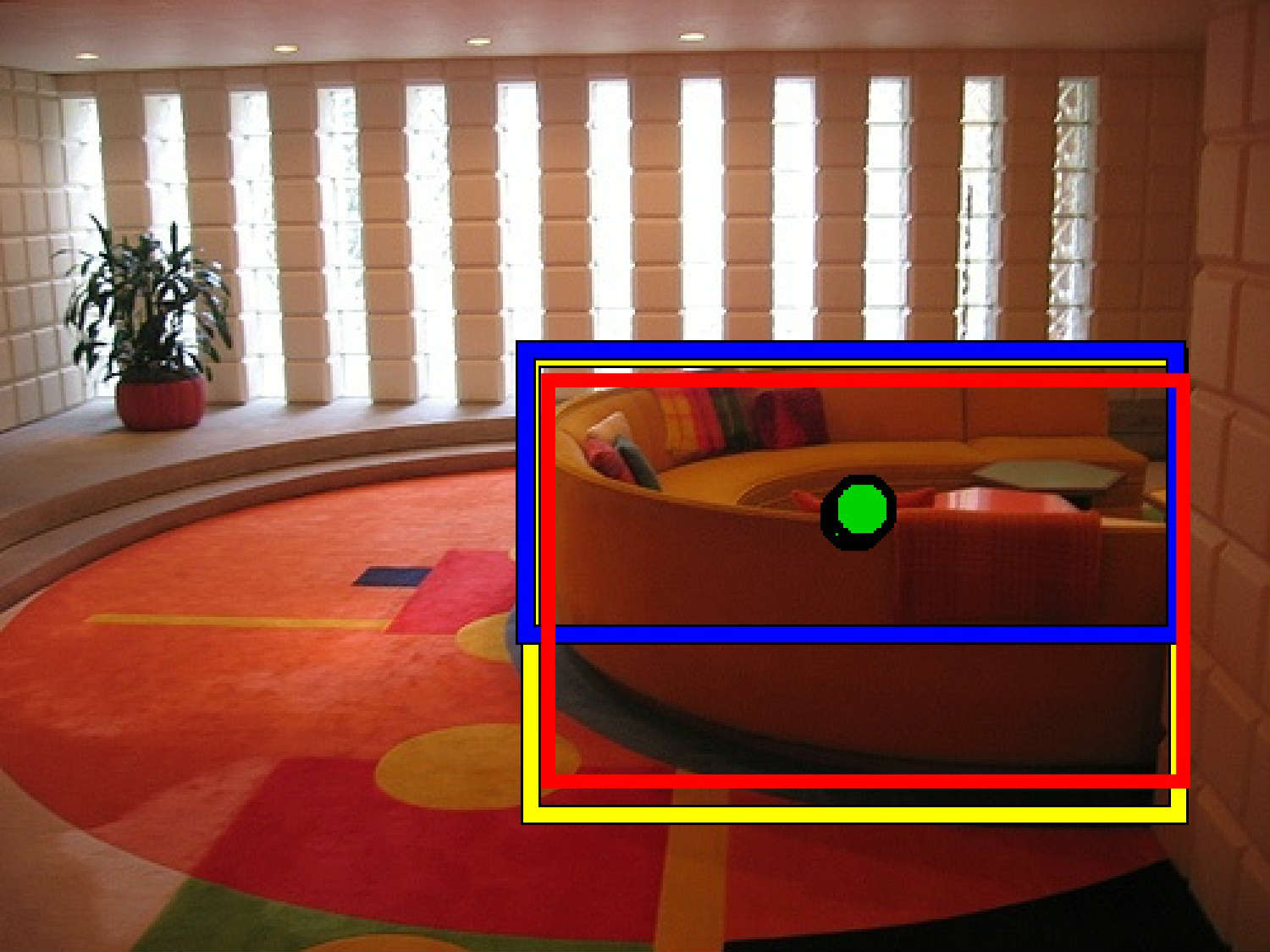}}\hfill
\subfloat{\includegraphics[width=0.12\textwidth,height=0.1\textwidth]{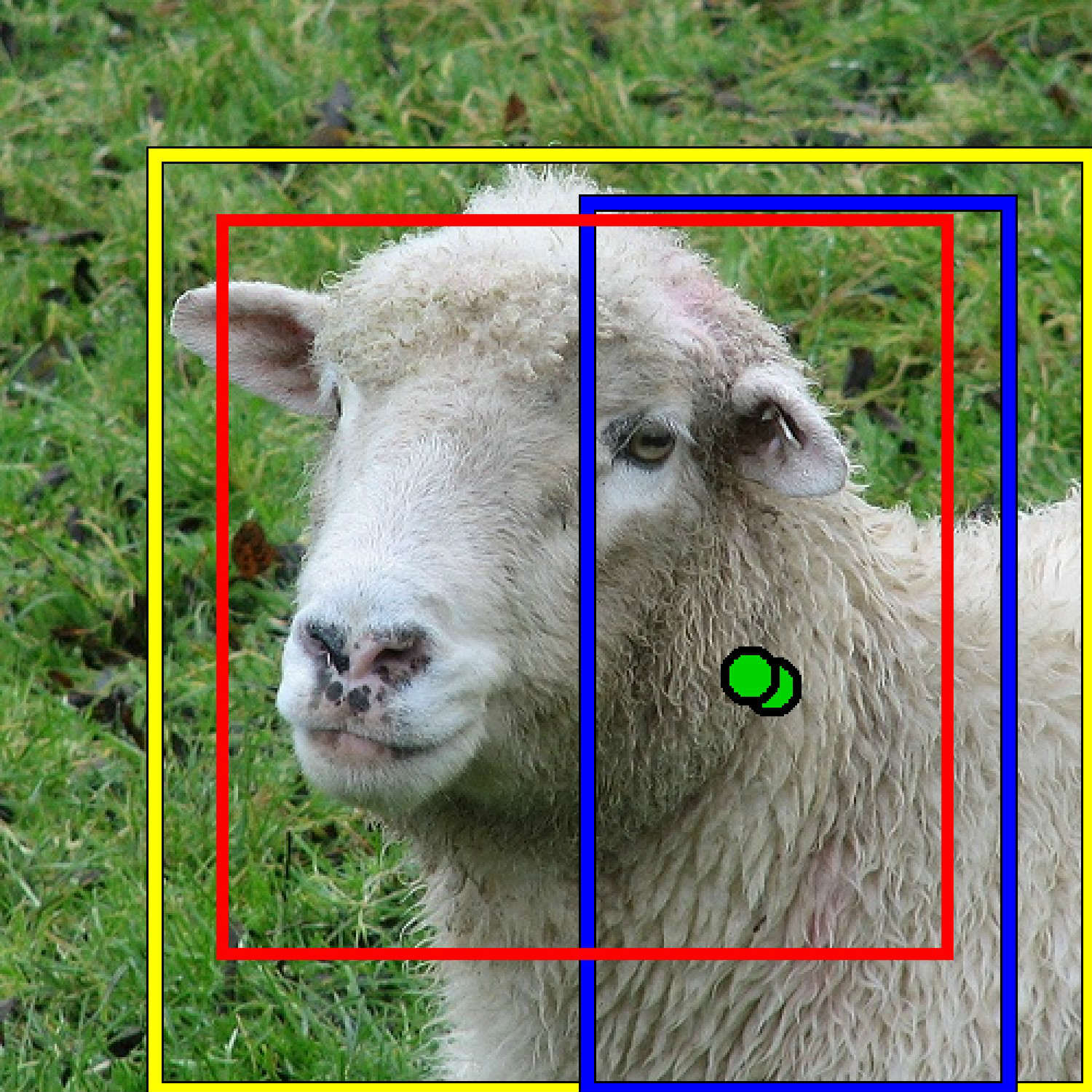}}\hfill
\subfloat{\includegraphics[width=0.12\textwidth,height=0.1\textwidth]{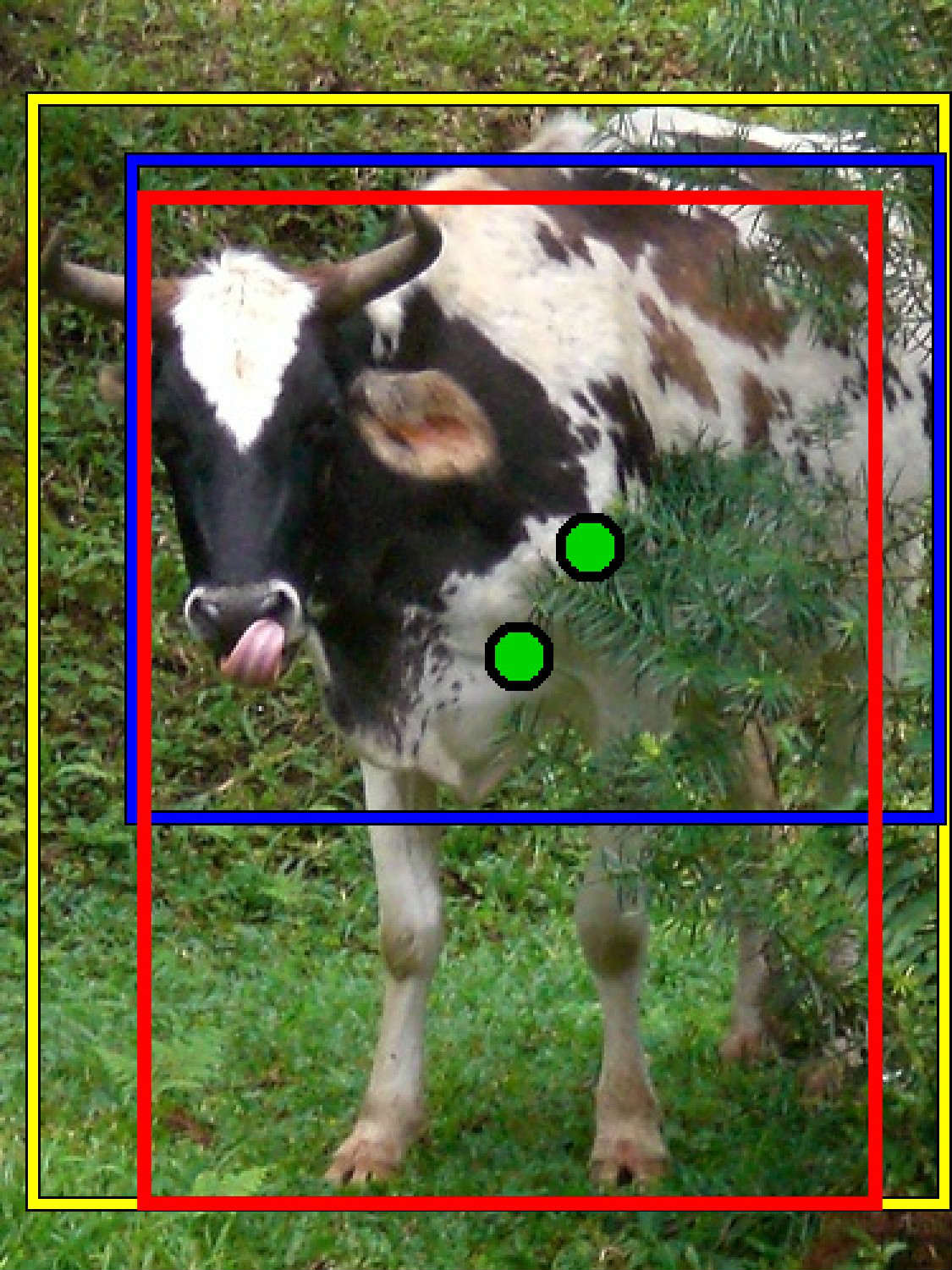}}\hfill
\caption{Examples of the pseudo ground-truths generated by our method and the center-click method on the PASCAL VOC2007 trainval set. The center-click method uses two clicks to generate the pseudo ground-truths (in green) while our method only uses one of them. For each example, the yellow bounding box is the ground-truth, and the blue one is the result obtained by the center-click method \cite{papadopoulos2017training} and the red one is the result obtained by our method.}
\label{examples}
\end{figure}

\textbf{Influence of SA-CAM.}
We conduct the experiments to verify the effectiveness of SA-CAM for generating the CAMs compared with Grad-CAM. As shown in Table \ref{tab:compare_cor_and_runtime}, the proposed SA-CAM based method achieves 79.6\% CorLoc while the Grad-CAM based method achieves 74.0\%, which is lower 
\begin{table}[!ht]
\centering
\caption{CorLoc (\%) and Running Time (FPS) Comparison Between Our Method Using SA-CAM and Grad-CAM on the PASCAL VOC2007 Trainval Set}
\label{tab:compare_cor_and_runtime}
\begin{tabular}{lcc}
\hline
Method &CorLoc & Time \\
\hline
Grad-CAM& 74.0 & 8.4\\
SA-CAM & 79.6 & 8.6\\
\hline
\end{tabular}
\end{table}
than the proposed SA-CAM based method. The reason is that SA-CAM is spatial-wise, which is more robust than the Grad-CAM (channel-wise). We also show the running time comparison for generating the CAMs in the Table \ref{tab:compare_cor_and_runtime}. The proposed SA-CAM based method runs at 8.6 fps, which is slightly faster than the Grad-CAM based method (8.4 fps).
\subsection{Qualitative Results}
 In Fig. \ref{examples}, we show some examples of the pseudo ground-truths generated by our method and the center-click based method \cite{papadopoulos2017training} compared with the ground-truths. The pseudo ground-truths generated by our method are more tight to cover the whole object regions than the center-click based method. 

\section{Conclusion}
In this paper, we propose a simple yet effective method by combining CNN visualization with center click to estimate the object scale for object detection. We use one click as supervision information, and with the help of the proposal selection algorithm and the proposed Spatial Attention (SA-CAM), we can estimate the object scale accurately and generate high-quality pseudo ground-truths. Experimental results show the superiority of the proposed method compared with the state-of-the-art methods.
\section{Acknowledgments}
This work is supported by the National Natural Science Foundation of China (Grant No. U1605252, 61872307 and 61571379) and the National Key Research and Development Program of China (Grant No. 2017YFB1302400).
\ifCLASSOPTIONcaptionsoff
  \newpage
\fi

\bibliographystyle{IEEEtran}

\begin{thebibliography}{1}
\bibitem{deng2009imagenet} J. Deng, W. Dong, R. Socher, L. J. Li, K. Li, and F. F. Li, ``Imagenet: A large-scale hierarchical image database," in \emph{IEEE Conf. Comput. Vis.  Pattern  Recognit}., 2009, pp. 248-255.
\bibitem{coco} T. Y. Lin, M. Maire, S. Belongie, J. Hays, P. Perona, D. Ramanan, P. Doll{\'a}r, and C. L. Zitnick, ``Microsoft coco: Common objects in context," in \emph{Eur. Conf. Comput. Vis.}, 2014, pp. 740--755.
\bibitem{ren2015faster} S. Ren, K.M. He, R. Girshick, and J Sun, ``Faster R-CNN: Towards Real-Time Object Detection with Region Proposal Networks," in \emph{Adv. in Neural Inf. Process. Syst}., 2015, pp. 91-99.
\bibitem{girshick2015fast} R. Girshick. ``Fast r-cnn," in \emph{IEEE Conf. Comput. Vis.  Pattern  Recognit}., 2015, pp. 1440-1448.
\bibitem{redmon2016you} J. Redmon, S. Divvala, R. Girshick, and A. Farhadi. ``You only look once: Unified, real-time object detection," in \emph{IEEE Conf. Comput. Vis.  Pattern  Recognit}., 2016, pp. 779-788.
\bibitem{liu2016ssd} W. Liu, D. Anguelov, D. Erhan, C. Szegedy, S. Reed, C.Y. Fu, and A. C. Berg, ``Ssd: Single shot multibox detector," in \emph{Eur. Conf. Comput. Vis}., 2016, pp. 21-37.
\bibitem{mod} J. L. Li, H. C. Wong, S. L. Lo, and Y. C. Xin, ``Multiple object detection by a deformable part-based model and an R-CNN," \emph{IEEE Signal Process. Lett}., vol. 25, no. 2, pp. 288-292, 2018.
\bibitem{bilen2016weakly} H. Bilen, A. Vedaldi, ``Weakly supervised deep detection networks," in \emph{IEEE Conf. Comput. Vis.  Pattern  Recognit}., 2016, pp. 2846-2854.
\bibitem{jie2017deep} Z. Q. Jie, Y. C. Wei, X. J. Jin, J. S. Feng, and W. Liu, ``Deep self-taught learning for weakly supervised object localization," in \emph{IEEE Conf. Comput. Vis.  Pattern  Recognit}., 2017, pp. 1377-1385.
\bibitem{diba2017weakly} A. Diba, V. Sharma, A. Pazandeh, H. Pirsiavash, and L. V. Gool, ``Weakly supervised cascaded convolutional networks," in \emph{IEEE Conf. Comput. Vis.  Pattern  Recognit}., 2017, pp. 914-922.
\bibitem{shen2018generative} Y. H Shen, R. R. Ji, S.C. Zhang, W. M Zuo, and Y. Wang, ``Generative adversarial learning towards fast weakly supervised detection," in \emph{IEEE Conf. Comput. Vis.  Pattern  Recognit}., 2018, pp. 5764-5773.
\bibitem{wang2018collaborative} J. J. Wang, J. C. Yao, Y. Zhang, and R. Zhang, ``Collaborative learning for weakly supervised object detection," in \emph{Int. Joint Conf. Artificial Intell}., 2018, pp. 971-977.
               
\bibitem{TS2C} Y. C. Wei, Z. Q. Shen, B. W. Cheng, H. H. Shi, J. J. Xiong, J.S. Feng, and T. Huang, ``Ts2c: Tight box mining with surrounding segmentation context for weakly supervised object detection," in \emph{Eur. Conf. Comput. Vis}., 2018, pp. 434-450.
\bibitem{pcl} P. Tang, X. G Wang, S. Bai, W. Shen, X. Bai, W. Y. Liu, and A. L. Yuille, ``Pcl: Proposal cluster learning for weakly supervised object detection," \emph{IEEE Trans. Pattern Anal. Mach. Intell} (Early Access)., 2018.
\bibitem{Ge_2018_CVPR_multi_evidience} W. F Ge, S. B. Yang, and Y. Z. Yu, ``Multi-evidence filtering and fusion for multi-label classification, object detection and semantic segmentation based on weakly supervised learning," in \emph{IEEE Conf. Comput. Vis.  Pattern  Recognit}., 2018, pp. 1277-1286.
\bibitem{tang2018weaklyrpn} P. Tang, X. G. Wang, A. T. Wang, Y. L. Yan, W. Y Liu, J. Z. Huang, and A. Yuille, ``Weakly supervised region proposal network and object detection," in \emph{Eur. Conf. Comput. Vis}., 2018, pp. 352-368.
\bibitem{papadopoulos2017training} D. P. Papadopoulos, J. R. R. Uijlings, F. Keller, and V. Ferrari, ``Training object class detectors with click supervision," in \emph{IEEE Conf. Comput. Vis.  Pattern  Recognit}., 2017, pp. 6374-6383.
\bibitem{zhou2016learningcam} B. L. Zhou, A. Khosla, A. Lapedriza, A. Oliva, and A. Torralba, ``Learning deep features for discriminative localization," in \emph{IEEE Conf. Comput. Vis.  Pattern  Recognit}., 2016, pp. 2921-2929.
\bibitem{selvaraju2017grad} R. R. Selvaraju, M. Cogswell, A. Das, R. Vedantam, D. Parikh, and D. Batra, ``Grad-cam: Visual explanations from deep networks via gradient-based localization," in \emph{IEEE Conf. Comput. Vis. Pattern  Recognit}., 2017, pp. 618-626.
\bibitem{li2018tellgain} K. P. Li, Z. Y. Wu, K. C. Peng, J. Ernst, and Y. Fu, ``Tell me where to look: Guided attention inference network," in \emph{IEEE Conf. Comput. Vis.  Pattern  Recognit}., 2018, pp. 9215-9223.
\bibitem{uijlings2013selective} J. R. Uijlings , K. E. V. D. Sande, T. Gevers, and A. W. Smeulders, ``Selective search for object recognition," \emph{Int. J. Comput. Vis}., vol.104, no. 2, pp. 154-171, 2013.
\bibitem{simonyan2014verydeep} K. Simonyan, and A. Zisserman, ``Very deep convolutional networks for large-scale image recognition," in \emph{Int. Conf. Learning Representations}., 2015, pp. 1-14.
\bibitem{everingham2010pascal} M. Everingham, L. V. Gool, C. K. Williams, J. Winn, and A. Zisserman, ``The pascal visual object classes (voc) challenge," \emph{Int. J. Comput. Vis}., vol.88, no. 2, pp. 303-338, 2010.
\bibitem{deselaers2012weakly} T. Deselaers, B. Alexe, and V Ferrari, ``Weakly supervised localization and learning with generic knowledge," \emph{Int. J. Comput. Vis}., vol.100,  no. 3, pp. 275-293, 2012.
\bibitem{jia2014caffe} Y. Q. Jia, E. Shelhamer, J. Donahue, S. Karayev, J. Long, R. Girshick, S. Guadarrama, and T. Darrell, ``Caffe: Convolutional architecture for fast feature embedding," In \emph{ACM Int. Conf. Multimedia}., 2014, pp. 675-678.
\bibitem{krizhevsky2012imagenet} A. Krizhevsky, I. Sutskever, and G. E. Hinton, ``Imagenet classification with deep convolutional neural networks," in \emph{Adv. in Neural Inf. Process. Syst}., 2012, pp. 1097-1105.
\end{thebibliography}
\end{document}